\documentclass[10pt,journal,compsoc]{IEEEtran}

\usepackage[nocompress]{cite}

\ifCLASSINFOpdf
  
\else

\fi
\usepackage{lipsum}
\usepackage{amsmath}
\usepackage{algorithm}
\usepackage[noend]{algpseudocode}
\usepackage{graphicx}
\usepackage{stfloats}
\usepackage{subfigure}
\usepackage{caption}
\usepackage{setspace}
\usepackage{booktabs}
\usepackage{multirow}
\usepackage{colortbl}
\definecolor{tabcolor}{rgb}{.105,.410,.113}
\usepackage{float}

\usepackage[colorlinks,
            linkcolor=black,
            anchorcolor=black,
            citecolor=black]{hyperref}

\usepackage{ragged2e}

\usepackage{xcolor}

\begin{document}

\title{SpaceEditing: Integrating Human Knowledge into Deep Neural Networks via Interactive Latent Space Editing}

\author{Jiafu~Wei, 
        Ding~Xia, 
        Haoran~Xie, 
        Chia-Ming~Chang, 
        Chuntao~Li, 
        and~Xi~Yang

\IEEEcompsocitemizethanks{\IEEEcompsocthanksitem%
Jiafu Wei, Chuntao Li, and Xi Yang (corresponding author) are with Jilin University. E-mail: weijf21@mails.jlu.edu.cn, \{lct33, yangxi21\}@jlu.edu.cn.
}

\IEEEcompsocitemizethanks{\IEEEcompsocthanksitem%
Ding Xia and Chia-Ming Chang are with the University of Tokyo. E-mail: dingxia1995@gmail.com, info@chiamingchang.com.
}

\IEEEcompsocitemizethanks{\IEEEcompsocthanksitem%
Haoran Xie is with Japan Advanced Institute of Science and Technology. E-mail: xie@jaist.ac.jp.
}}

\markboth{Journal of \LaTeX\ Class Files,~Vol.~14, No.~8, August~2015}%
{Shell \MakeLowercase{\textit{et al.}}: Bare Demo of IEEEtran.cls for Computer Society Journals}

\IEEEtitleabstractindextext{

\begin{center}
\setcounter{figure}{0}
\includegraphics[width=\linewidth]{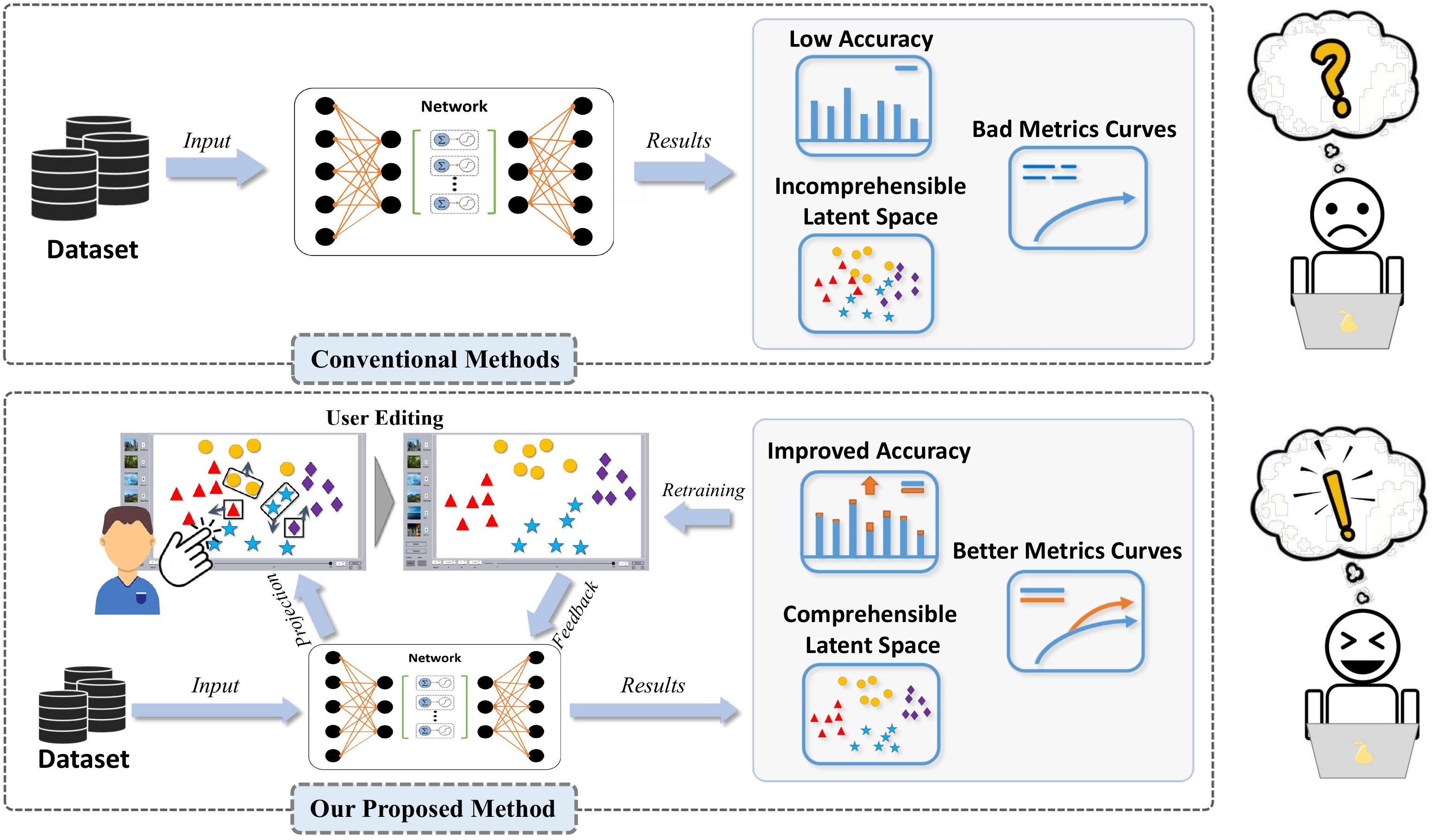}
\captionof{figure}{Using conventional methods to train the network, the training process of the network is a black box, and the training process is uncontrollable. Our proposed method projects the high-dimensional features of the data into a 2D workspace, where the user can manually edit the high-dimensional features. Using this method to train the network can control the training process of the network to a certain extent, which not only allows people to better understand the training process of the network but also integrates human knowledge into the training process of the network, thereby improving the performance of the network.}
\label{fig:figure1}
\end{center}

\justify 
\begin{abstract}
We propose an interactive editing method that allows humans to help deep neural networks (DNNs) learn a latent space more consistent with human knowledge, thereby improving classification accuracy on indistinguishable ambiguous data. Firstly, we visualize high-dimensional data features through dimensionality reduction methods and design an interactive system \textit{SpaceEditing} to display the visualized data. \textit{SpaceEditing} provides a 2D workspace based on the idea of spatial layout. In this workspace, the user can move the projection data in it according to the system guidance. Then, \textit{SpaceEditing} will find the corresponding high-dimensional features according to the projection data moved by the user, and feed the high-dimensional features back to the network for retraining, therefore achieving the purpose of interactively modifying the high-dimensional latent space for the user. Secondly, to more rationally incorporate human knowledge into the training process of neural networks, we design a new loss function that enables the network to learn user-modified information. Finally, We demonstrate how \textit{SpaceEditing} meets user needs through three case studies while evaluating our proposed new method, and the results confirm the effectiveness of our method.
\end{abstract}


}

\maketitle


\IEEEpeerreviewmaketitle

\section{Introduction}

\IEEEPARstart{A}{lthough} deep neural networks (DNNs) maintain excellent results in classification, they still struggle to distinguish similar ambiguous data. The machine learning community has realized the disadvantage of networks in dealing with abstract things \cite{r4001}, such as things like shapes and concepts. In addition, when faced with datasets in some specialized fields, such as archaeological-related datasets, the performance of the network is not satisfactory. The domain dataset requires corresponding domain knowledge. Therefore, the recognition of features by humans should help the learning of deep learning networks.

However, the learning process of current deep learning networks is still uncontrollable (Fig. \ref{fig:figure1}). In general, whether in the training process or the fine-tuning process, people can only judge the effect of network training through the results of loss or metrics, but cannot directly process the data. If people want to achieve better results, they can only start with traditional fine-tuning methods such as adjusting hyperparameters, but such methods often require multiple debugging to achieve better results. To let people participate in the network training process more intuitively, the role of high-dimensional features in network training has been paid more and more attention\cite{r355}. High-dimensional features can be observed through projection visualization \cite{r302,r313,r314}, but existing methods cannot directly affect high-dimensional features, and the latent space learned by existing deep learning networks is still uncontrollable.

Interactive machine learning is a good direction to address the above problem, which tries to let human knowledge help the network learn. For example, Sakata et al. \cite{r111crownn} introduce a network called CROWNN, which allows people to participate in the network classification process, and then it leverages the learned human strengths to better perform classification tasks.

The user interface is one of the important means to support interactive machine learning. On the one hand, the user interface can help people observe the relationship between data more intuitively. On the other hand, the user can participate in the training of the network through the interactive function of the user interface. Therefore, it is necessary to design a suitable spatial layout structure for the user interface. For example, Chang et al. \cite{ming} used the spatial layout idea to design a system for improving the annotation quality of non-professional annotators.

\begin{figure*}[h]
\centering
\includegraphics[width=1.00\textwidth]{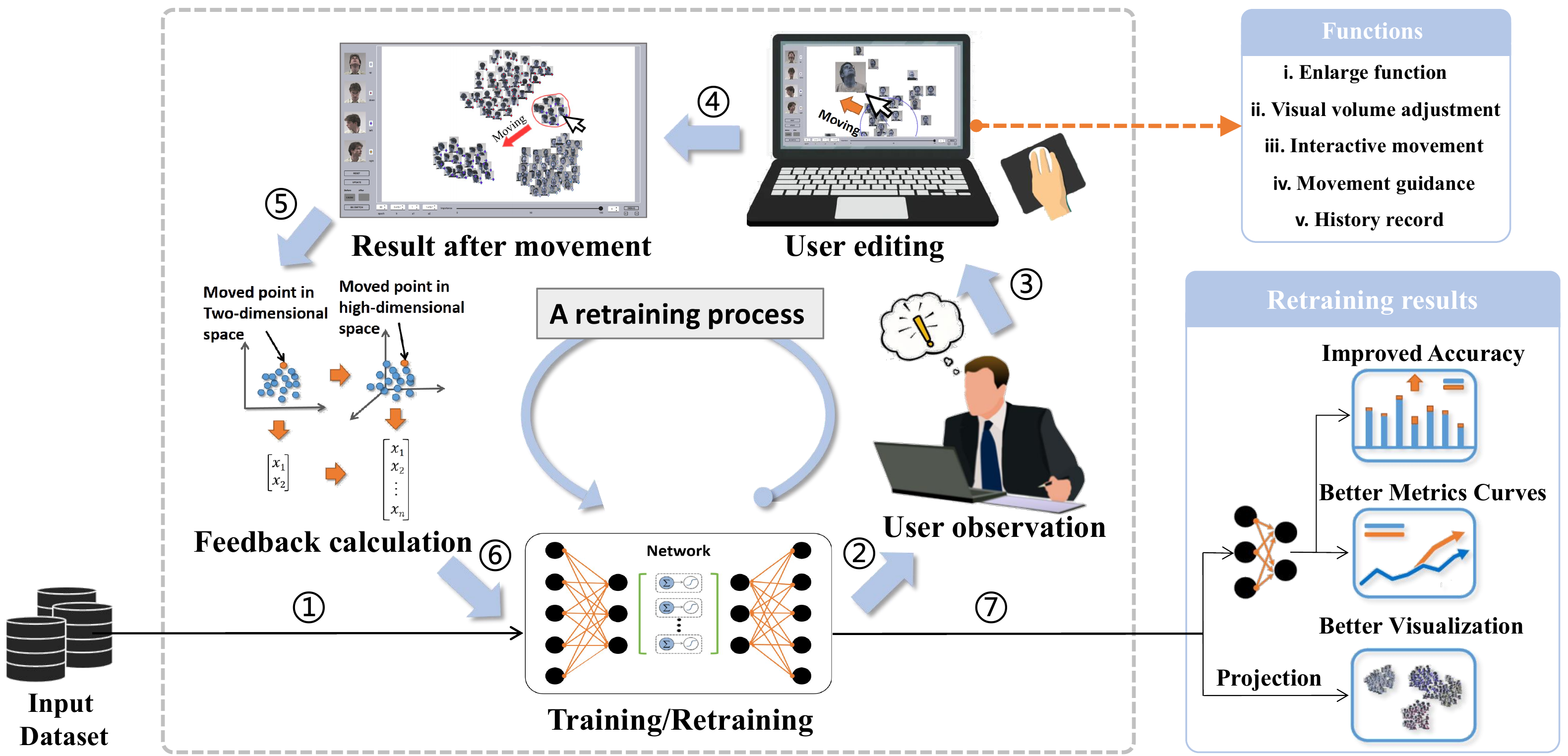}
\renewcommand*{\thefigure}{2}
\caption{\label{fig:figure2} The workflow of our proposed method. (1) In the preprocessing stage, the raw data is trained to obtain a network for visualizing the images. (2) The data features output by the network are projected into a 2D workspace for the user to observe. (3) In the 2D workspace, users can easily observe the visualized data, and at the same time, they can use the functions provided by the system to easily interact with the projection points. (4) When the user edits the projection point in the 2D workspace, (5) the system automatically performs calculations in a high-dimensional latent space. (6) The network will learn according to the changes of projection points before and after moving, (7) therefore achieving the purpose of the user helping the network to retrain.}
\end{figure*}

Although in past work, we proposed a method to fine-tune the network using 2D projections of the data\cite{rwei}, this method does not work directly on high-dimensional features, and at the same time, the effect of this method is not significant. Therefore, in this paper, we propose a novel interactive machine learning method to address the above issues. Through our designed system, users can intuitively observe the distribution of data in latent space. Then, users can also modify the high-dimensional features in the latent space, and the modified information will be used to retrain the network to improve the performance of the network. Our method realizes the visualization of high-dimensional features by projecting the high-dimensional features in the network onto a 2D workspace, and the location information of the projected point reflects the classification result of the point to a certain extent. Based on the various interactive functions of our system, users can reclassify the projection results based on their knowledge, and can also move the points they think are misclassified to the positions they think are correctly classified. Then, the system can automatically feed back the result of the user's movement to the network for retraining (Fig. \ref{fig:figure2}). 

    

To summarize, our contributions include:

\begin{enumerate}
   
    \item We propose a novel and effective method that enables users to interactively edit the latent space based on their knowledge, thereby guiding the network's learning process. This method can incorporate human knowledge into the training process of the network, which not only makes the network jump out of the local minimum area and improves the performance of the network but also allows users to obtain a more understandable latent space.
    \item We design a new interactive system, \textit{SpaceEditing}, which can apply our proposed method and allow users to synchronize their operations from two-dimensional space to high-dimensional space. In addition, it also has various interactive functions such as enlarge function, visual volume adjustment, interactive movement, movement guidance, and history record, which provide feasibility for manual editing of latent space.
    \item We conduct three case studies to evaluate the effect of \textit{SpaceEditing} on different types of machine learning tasks and users with different identities. The case studies indicate: the usefulness of the system; the effectiveness of the proposed method; the flexibility to adapt to different scenarios; the user's experience and evaluation.

\end{enumerate}

\section{Related work}


\subsection{Interactive Machine Learning}
With the rapid development of graphical interfaces, the importance of humans in the working and learning process of machines has become increasingly apparent \cite{r122,r123,r124}. The advantage of IML (Interactive Machine Learning) is that the addition of humans can help machines complete abstract tasks that are difficult for machines \cite{r111crownn,r115,r117,r17,r32}. Machine learning can be used as a creative tool, and human participation can help group machine learning in a creative and artistic direction \cite{r34,r39}. Through an interactive method, it is easier for the machine to generate the data and networks the user wants \cite{r2020,r2021,r119,r117}. In addition, some studies add brain interface to the process of machine learning \cite{r113}, which not only makes the system show stronger learning ability \cite{r115} but also shortens the training time of the network \cite{r112}.

Furthermore, IML systems pay special attention to user experience and user understanding of the system due to human involvement, so it is necessary to design robust and user-friendly systems \cite{r16,r62,r61,r2010}. Human-oriented is the premise to achieve this goal. Driven by the human-oriented design concept\cite{r37,r116,r2028,r35}, many meaningful new methods of interactive machine learning have been born\cite{r2032,r2033}.

There are many works based on the idea of IML, but there is no way to train the network through human interaction with the latent space, and the potential of the latent space has not been tapped. To this end, we propose a new deep learning method for interacting with the latent space.

\subsection{Conventional Fine-tuning Methodes}

Fine-tuning is an important method in the field of machine learning. The time required to train a new network can be greatly simplified by modifying part of the network to the network required by the user \cite{r206}. Based on traditional fine-tuning methods, many novel fine-tuning methods have been generated. For example, by changing part of the network structure to retrieve and classify data such as images, Radenovi{\'c} et al. \cite{r208} finally realized a fine-tuning method without manual annotation. Rosa et al. \cite{r207} introduce harmony search and some of its variants to fine-tune image classification, filling a gap in CNN parameter optimization research. Observing that not all parameters need to be updated during fine-tuning, Xu et al. \cite{r212} proposed an efficient fine-tuning technique CHILD-TUNING, which masks the gradients of non-sub-networks in the reverse process. In addition to the above methods, visualization techniques are also applied to fine-tune the network. For example, Amershi et al. \cite{r200} designed a visualization tool ModelTracker, which can analyze the performance of the network and help users fine-tune the network.

Latent space plays a very important role in network training \cite{r355}. Although the above works have achieved very good results in network fine-tuning, they have not proposed new fine-tuning methods from the perspective of latent space. In addition, in the datasets of some specialized fields, it is usually necessary to combine a large amount of domain knowledge to achieve better results \cite{rexpert}, therefore the human-in-the-loop method is very suitable. For example, to address the low accuracy of medical image segmentation, Wang et al. \cite{rinteractivatefine-tuning} proposed a novel interactive segmentation framework. Based on the above ideas, we propose a novel retraining method from latent space and human-in-the-loop perspectives.

\subsection{Spatial Layout}
The concept of spatial layout is widely used in management because spatial layout not only stores information but also reflects the relationship between information to a certain extent \cite{r5001,r5002}. There are many meaningful types of research based on spatial layouts, such as dimensionality reduction visualization \cite{r5003,r5004}. By mapping high-dimensional data to a 2D or 3D interface, people can more easily observe the layout relationship between the data and then analyze the data \cite{r301,r302,r313,rmds}. The spatial layout can also reflect the relationship between icons, objects, and other information \cite{r306,r312}. Reasonable use of spatial layout can achieve the purpose of assisting users. For example, to address the lack of professional annotators, Chang et al. \cite{ming} utilize spatial layout to design a novel annotation interface to improve the annotation quality of non-expert image annotations. Wang et al. \cite{r311} used the hierarchical spatial structure to optimize the way the map, which solved the tedious process of users zooming in to see map details and zooming out to see an overview. Mai et al. \cite{r315} discussed and studied the factors affecting the user's spatial layout, and showed the factors affecting the user's related spatial layout. By combining the advantages of small multiples and visual aggregation with interactive browsing, Lekschas et al. \cite{r2016} propose a structured design space to guide the design of visual-spatial layouts.

Numerous studies have proven that information can be effectively managed through spatial arrangement \cite{r3001,r307,r2030}. One example is a spatial search system that makes it easy for users to search for desired information in 2D space by interacting with the visualized data \cite{r303}. Chen et al. \cite{r304} designed a system to facilitate bug discovery through semantic data search, in which users interactively create a topology to convey information in a spatial layout. Human-centric spatial layout techniques have also emerged, which visualize time-series data through user interfaces to enhance human-to-human collaboration \cite{r308}. Asai et al. \cite{r309} combine a code editor with an interactive scatterplot editor enabling users to effectively understand the behavior of statistical modeling algorithms. With the continuous development of technology, spatial layout, and machine learning have been continuously integrated, resulting in a large number of novel and excellent research results \cite{r4002,r4003}. For example, Eisenstadt et al. \cite{r4004} leveraged machine learning techniques to process information representations about building room types and the spatial layout of individual rooms.

\section{System}





The latent space means the representation of encoded data, however it is incomprehensible in most cases. To further identify and understand the latent space, and integrate human capabilities into the machine training process, we design a novel interactive system \textit{SpaceEditing} that allows users to adjust the position of vectors in the latent space. Then, the system will retrain the parameters of the network with moved vectors to improve the performance step by step.

\subsection{Design Goals}



The primary design goal of \textit{SpaceEditing} is to provide a novel method to interact with the latent space for users. 
The system should highlight the relationships and connections between different types of data.
At the same time, the basic requirement is that the user can visually observe the distribution of the data in the latent space, and the user can easily interact with the vectors in the latent space. 





We proposed three primary designing goals for \textit{SpaceEditing}:
\begin{enumerate}
    \item \textbf{Visualize the data representation intuitively.} A fundamental function is the visualization of the latent space, where our system could clearly and correctly describe the similarities and differences within data representations. In this way, users can locate target data and make the right modification without effort.
    \item \textbf{Design an effective interactive system.} A batch-selection mechanism is critical for a dataset comprising millions of images. With this kind of function, we can alleviate the burden that users need to process tons of data one by one. Besides, to assist the batch-selection mechanism, we need to develop a set of interaction mechanisms accordingly.
    \item \textbf{Facilitate users by highlighting ambiguous data.} A suggestion mechanism for ambiguous data (data with wrong predicted values and data in unreasonable locations) can substantially improve user experience. Therefore, we provide corresponding movement guidance functions for users. 
\end{enumerate}



\subsection{System Description}

\begin{figure*}
\centering
\includegraphics[width=1.00\textwidth]{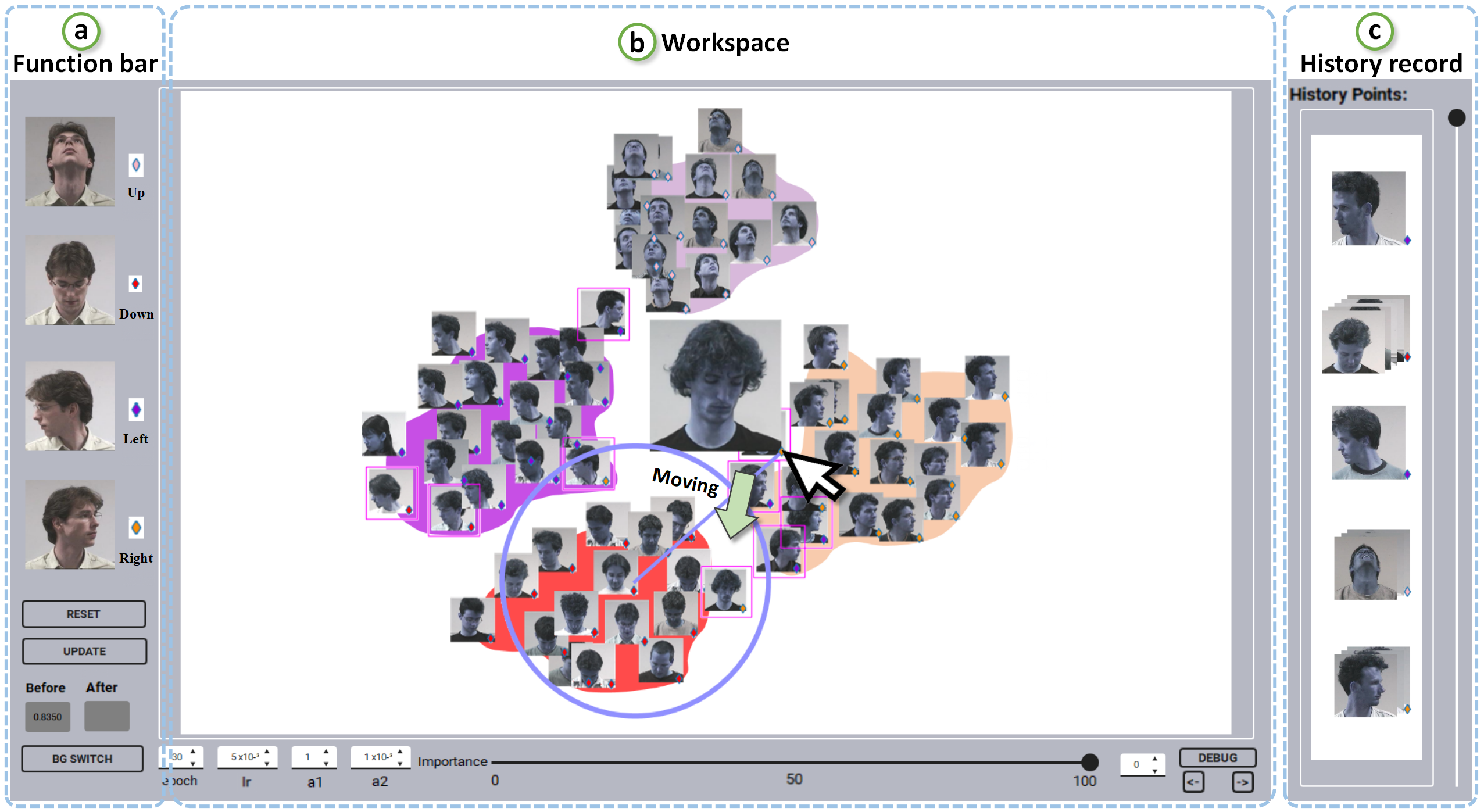}
\renewcommand*{\thefigure}{3}
\caption{\label{fig:figure4} System. The system can be divided into three parts: a) function bar, b) workspace and c) history record module. In the workspace, projected coordinates are combined with the data image itself. When the user mouses over a certain point, an enlarged image of the point will be displayed, and at the same time, a light purple guide line and guide circle will also be displayed to provide a general direction for the user to move. The workspace shows a total of four classes, corresponding to four colors. The history record module is used to store the points operated by the user.}
\end{figure*}

Our system consists of three parts: function bar, workspace, and history record (Fig. \ref{fig:figure4}).

\subsubsection{Function Bar}

\begin{figure}
\centering
\includegraphics[width=0.5\textwidth]{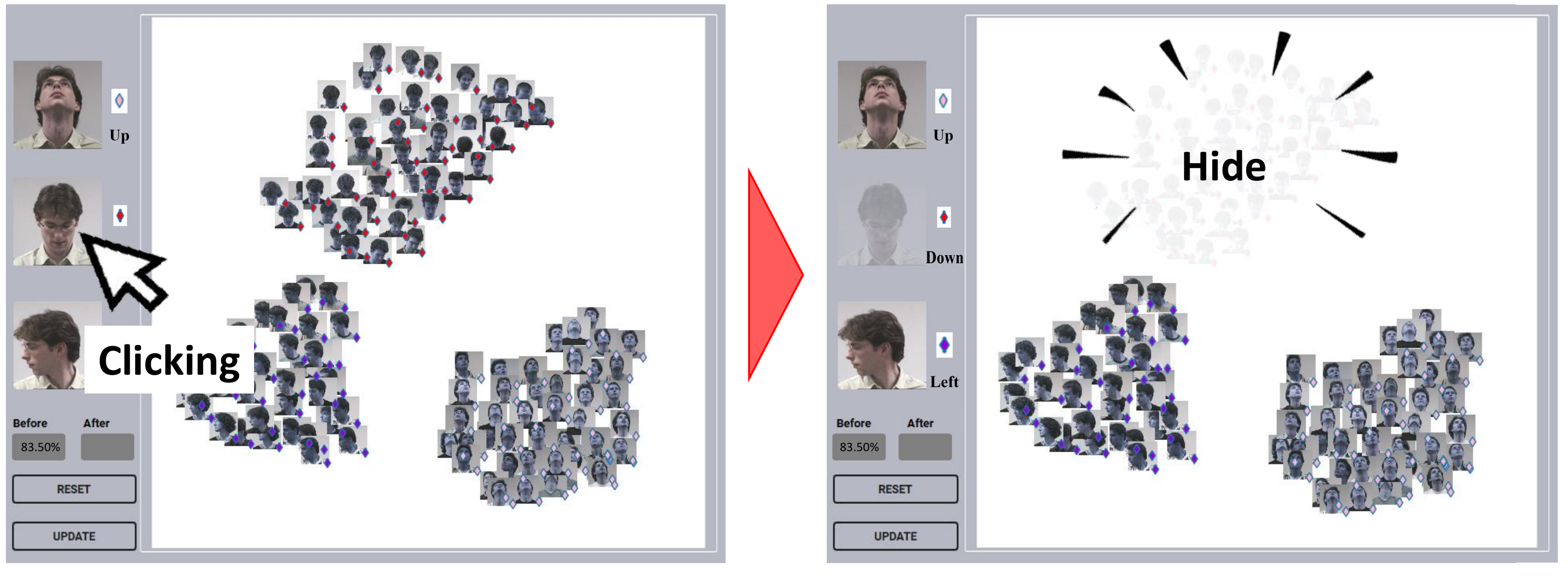}
\renewcommand*{\thefigure}{4}
\caption{\label{fig:figure82} Display and hide. Click the image representing the class on the left to hide or display the corresponding class in the workspace.}
\end{figure}

Data classes and representative images are displayed in the function bar. When the user clicks on the corresponding representative picture, the data of this class in the workspace can be controlled to be displayed or hidden, and the user can hide irrelevant data through this function (Fig. \ref{fig:figure82}).

There are two buttons below, the reset button is used to restore user operations, and the update button is used to control the network for retraining. The bottom display boxes show the accuracy before and after the update.

\subsubsection{Interactive User Workspace}

The system uses a novel spatial layout that helps users obtain a more comprehensible representation of the latent space (Fig. \ref{fig:figure4}(b)) . We apply Isomap \cite{risomap} to latent representations of the data, resulting in a 2D visual layout, the reason for applying Isomap as a dimensionality reduction method is explained in Section 3.4. In order to display the data more intuitively on the workspace, the data itself is displayed in the form of a thumbnail, and the predicted class of the data is displayed in the form of a colored coordinate point in the lower right corner of the thumbnail, where different colors represent different predicted values. Through such a spatial layout design, not only the prediction of the data is clear at a glance, but also the data itself can be better presented to the user, which is convenient for the user to compare and interact with the data.


The distribution of the data reflects the classification results of the network. We used heatmaps of different colors as backgrounds for different classes, which play the role of guidance for user movement. In the workspace, users can explore and interactively modify the data in it. At the same time, this system also has basic redo and undo functions, which brings convenience to user operations.



Then enter the description of the main functions of the system.

\textbf{Enlarge function.}
How to facilitate users to compare data is an important consideration when we design the system. When the mouse hovers over a point, the image represented by the point is enlarged (Fig. \ref{fig:figure4}(b)). When the user is faced with the situation of distinguishing different data, this function can help the user distinguish mixed data, which is convenient for users to compare adjacent data and make better movement judgments. At the same time, the system also has basic interactive zoom and wheel pan functions.

\begin{figure}[h]
\centering
\includegraphics[width=0.5\textwidth]{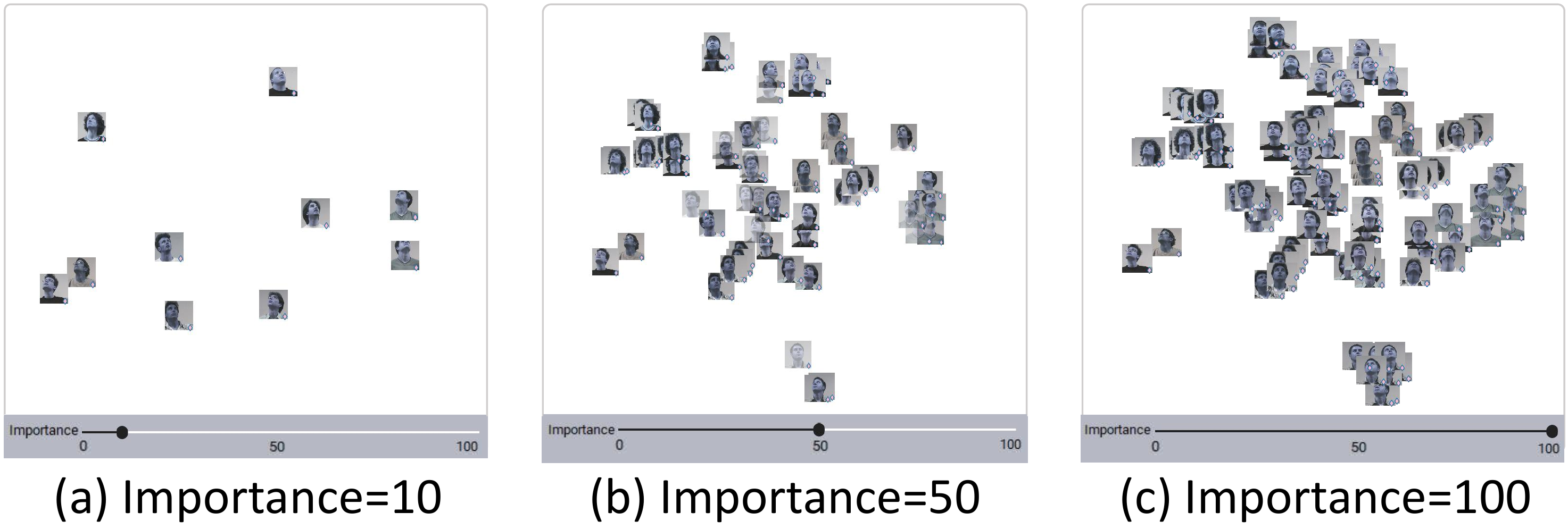}
\renewcommand*{\thefigure}{5}
\caption{\label{fig:figure83}Visual volume adjustment. An example of using the importance slider bar to adjust the amount of displayed data.}
\end{figure}

\textbf{Visual volume adjustment.}
If the amount of data is too large, it will inevitably affect the user's observation, therefore the system adds the function of visual volume adjustment. The user can control the data range displayed on the current screen by dragging the slider bar below the workspace (Fig. \ref{fig:figure83}). The slider bar indicates the importance of the data from left to right. We define importance as how confident the network is about the outputs. Therefore, we use the softmax function to sort the maximum value of the network output in descending order, and the sorted result is the importance we define. We associate importance with the range of data displayed. In addition, the user can also control the number of data displayed on the current screen through the input window on the right side of the slider bar. After entering a number, the workspace will only display the corresponding number of projected points. Users can hide over-displayed data according to the above functions.

\begin{figure}[h]
\centering
\includegraphics[width=0.5\textwidth]{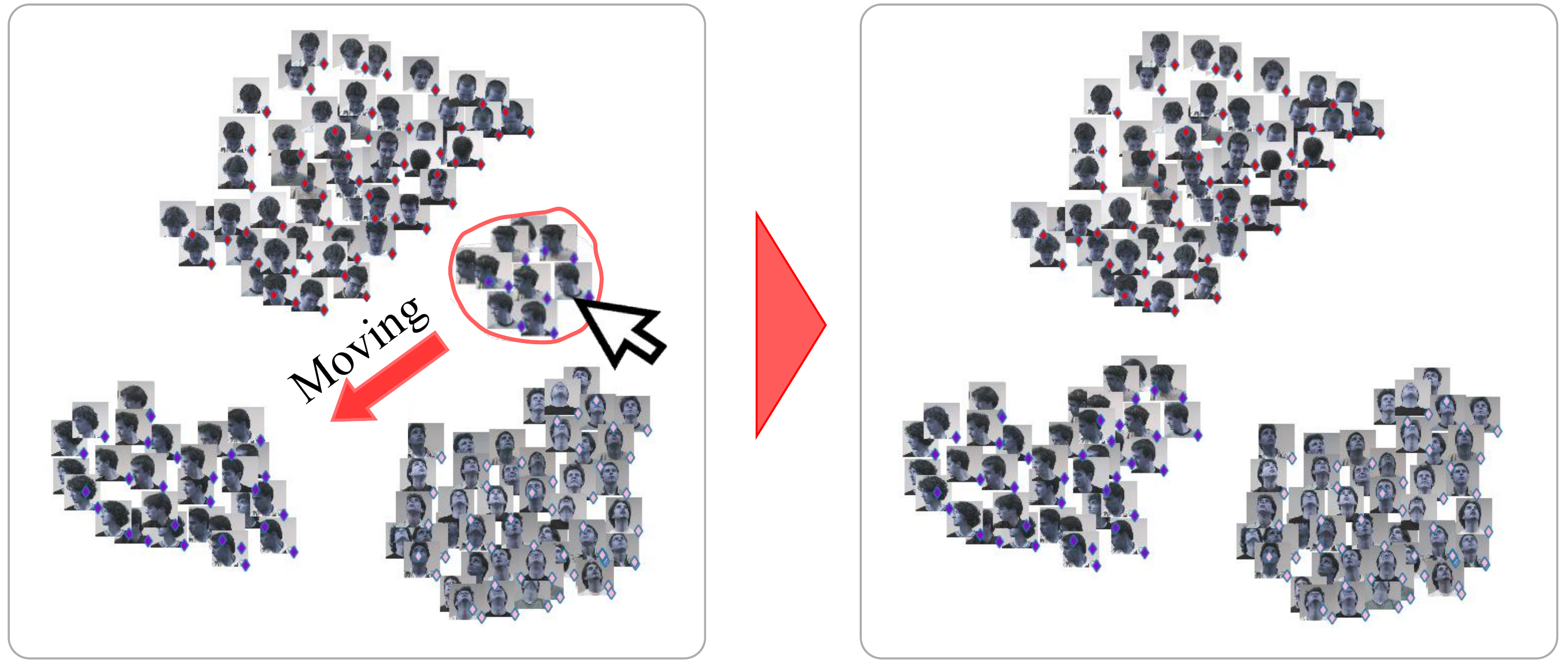}
\renewcommand*{\thefigure}{6}
\caption{\label{fig:figure81}Interactive movement. An example of multi-point movement using the Lasso tool.}
\end{figure}

\textbf{Interactive movement.}
The most basic interactive function in the system is to move the projected coordinates in the user's workspace. The user can move a single point, or use the Lasso tool to move multiple points at the same time (Fig. \ref{fig:figure81}).

\textbf{Movement guidance.}
To greatly facilitate the user's operation, the system should have some guidance functions. Our system uses a pink box to highlight data with incorrect predictions to facilitate the user to select the data to be moved (Fig. \ref{fig:figure4}(b)). In addition, the system adds a guideline and a guide circle to remind the user of the approximate location of the cluster formed by the ground truth of the current moving point. We also use heatmaps to show the distribution of each class, which can also be used as a reference for user movement.


\subsubsection{History Record}
During the interaction process, it is inevitable for users to make movement errors and want to modify the movement process. At the same time, users may also find interesting patterns and want to record their movements. Therefore, we provide the history record module (Fig. \ref{fig:figure4}(c)). The history of user movement will be stored in the history record module. By clicking on the corresponding history record, the workspace can be traced back to the corresponding state. Users can also use the display mechanism of the history module to make further comparisons of the data. In addition, this module provides redo and undo functions.

\subsection{Feedback Calculation}

\begin{figure}
\centering
\includegraphics[width=0.48\textwidth]{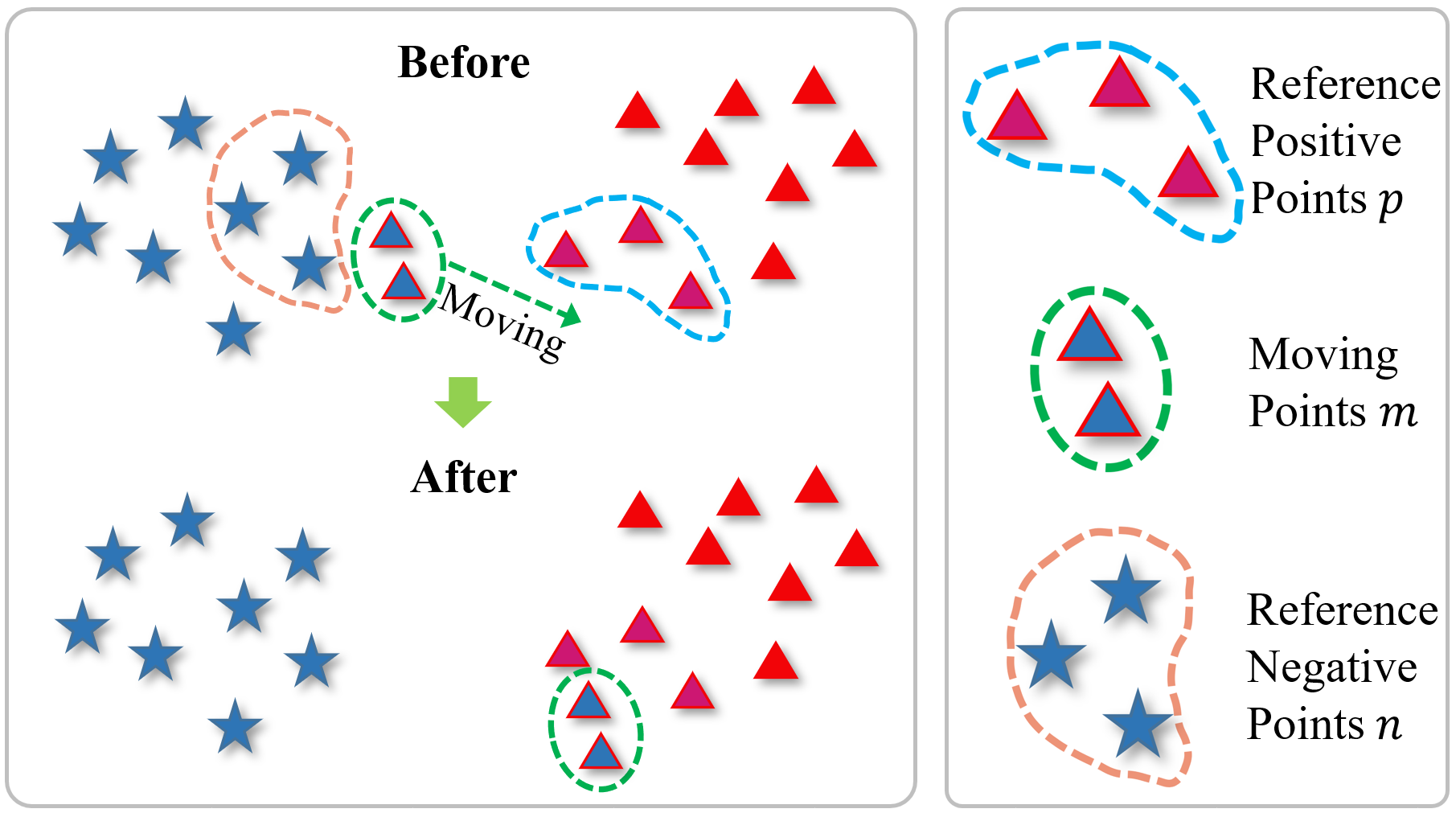}
\renewcommand*{\thefigure}{7}
\caption{\label{fig:figure3} An example of simulating user movement. The user first observes the data using the system functions, and finds the moving points \begin{math}\mathit{m}\end{math}. Then, the user moves the moving points \begin{math}\mathit{m}\end{math} to the place he thinks is appropriate according to his knowledge. The system will select reference positive points \begin{math}\mathit{p}\end{math} and reference negative points \begin{math}\mathit{n}\end{math} according to the position before and after the user moves.}

\end{figure}

In order to incorporate human knowledge into the training process of neural networks, we design a new loss function to allow the user's operations in the 2D workspace will be learned by high-dimensional hidden vectors, and the network can learn a latent space that is more in line with human knowledge.

We strive to make the image features of the user's movement in the latent space closer to the position after the user's movement in the latent space, for which we draw on the idea of triplet loss \cite{r3002}. The purpose of triplet loss is to make the features of the same label as close as possible in spatial position, while the features of different labels are as far away as possible in spatial position. Our purpose is to make the user's moving features closer to the moved spatial positions, therefore we need to find a high-dimensional feature \begin{math}\mathit{P}\end{math} in the high-dimensional space as a reference. In the 2D workspace, we set the point the user moved to point \begin{math}\mathit{m}\end{math}. We set the \begin{math}\mathit{k}\end{math} points with the same label closest to point \begin{math}\mathit{m}\end{math} after the user moves to point \begin{math}\mathit{p_i}\end{math} (Fig. \ref{fig:figure3}), where 0$<$\begin{math}\mathit{i}\end{math}$<$\begin{math}\mathit{k}\end{math}. To calculate the high-dimensional feature \begin{math}\mathit{P}\end{math} in the high-dimensional space, we perform a weighted sum over the points \begin{math}\mathit{pi}\end{math} (0$<$\begin{math}\mathit{i}\end{math}$<$\begin{math}\mathit{k}\end{math}), where the weight is the inverse of the distance between point \begin{math}\mathit{m}\end{math} and point \begin{math}\mathit{p_i}\end{math} in the high-dimensional space (see Formula (\ref{equ:p})).

\begin{equation}
\label{equ:p}
P = \sum\nolimits_{i}^k  \frac{1}{||m - p_i||_2^2} \times p_i
\end{equation}

where \begin{math}\mathit{m}\end{math} and \begin{math}\mathit{p}\end{math}$_{i}$ represent features in the high-dimensional space, corresponding to points found in two-dimensional space.

At the same time, in order to suppress the aggregation phenomenon of different classes of features to a certain extent, we define a high-dimensional feature \begin{math}\mathit{N}\end{math}. We set the \begin{math}\mathit{k}\end{math} points with different labels closest to point \begin{math}\mathit{m}\end{math} before moving to point \begin{math}\mathit{n_i}\end{math}, where 0$<$\begin{math}\mathit{i}\end{math}$<$\begin{math}\mathit{k}\end{math}. In the same way, the corresponding point \begin{math}\mathit{N}\end{math} is obtained according to point \begin{math}\mathit{n_i}\end{math} (0$<$\begin{math}\mathit{i}\end{math}$<$\begin{math}\mathit{k}\end{math}) (see Formula (\ref{equ:n})). What the loss function needs to do is to move the point \begin{math}\mathit{m}\end{math} in the high-dimensional space closer to the high-dimensional feature \begin{math}\mathit{P}\end{math}, and make the distance between point \begin{math}\mathit{m}\end{math} and high-dimensional feature \begin{math}\mathit{N}\end{math} in the high-dimensional space further apart, therefore the network can learn a latent space that is closer to the user’s movement.

\begin{equation}
\label{equ:n}
N = \sum\nolimits_{i}^k  \frac{1}{||m - n_i||_2^2} \times n_i
\end{equation}

where \begin{math}\mathit{n}\end{math}$_{i}$ represent features in the high-dimensional space, corresponding to points found in 2D space.

The classification loss \begin{math}\mathit{loss}\end{math}$_{cls}$ is obtained by performing a Cross-entropy (CE) calculation between the predicted labels and the ground-truth labels. The distance difference loss \begin{math}\mathit{loss}\end{math}$_{dis}$ is calculated based on the points the user moves (see Formula (\ref{equ:loss1})).

\begin{equation}
\label{equ:loss1}
loss_{dis} = \sum\nolimits_{i}^D max(||m_i - P_i||_2^2 - ||m_i - N_i||_2^2 + \delta , 0 )
\end{equation}

where \begin{math}\mathit{D}\end{math} represents the number of user moving points, \begin{math}\mathit{m}\end{math}$_{i}$, \begin{math}\mathit{P}\end{math}$_{i}$ and \begin{math}\mathit{N}\end{math}$_{i}$ represent corresponding features in high-dimensional space, and \begin{math}\mathit{\delta}\end{math} is the margin used to control the distance between points \begin{math}\mathit{P}\end{math}$_{i}$ and \begin{math}\mathit{N}\end{math}$_{i}$.

The total loss function is weighted by the classification loss and the distance difference loss before and after adjustment (see Formula (\ref{equ:loss3})). Using the total loss function, the network can be retrained based on where the user moves, and the output will be closer to what the user changed, therefore the user can assist the network in better classification.

\begin{equation}
\label{equ:loss3}
Loss = w_{cls}\times loss_{cls}  + w_{dis}\times loss_{dis}
\end{equation}



\subsection{Network and Dimensionality Reduction Method Selection}

After comparisons, we choose the classic ResNet18\cite{rresnet} as the basic classification network. We used the pre-trained network of ResNet18, and to avoid overfitting, we froze the first few layers of the network. The fully connected layer of the network consists of two layers, and the number of nodes in each layer is 2048 and 512 respectively. We apply a dimensionality reduction method on layer 512 to visualize the data on the user workspace. The network was trained with a learning rate between 0.001 and 0.0005 and used the Adam optimizer with a fixed batch size of 128.

\begin{figure}[h]
\centering
\includegraphics[width=0.45\textwidth]{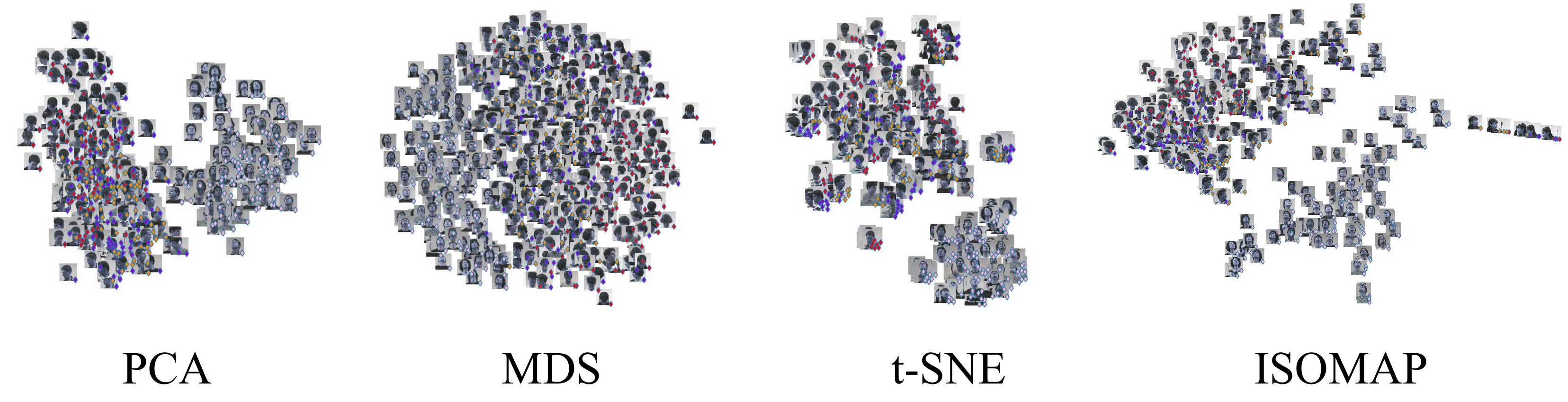}
\renewcommand*{\thefigure}{8}
\caption{\label{fig:figure116} Example of dimensionality reduction results for PCA, MDS, t-SNE, and Isomap.}

\end{figure}

We tested four commonly used dimensionality reduction methods on the garbage classification dataset, namely PCA, MDS, t-SNE, and Isomap (Fig. \ref{fig:figure116}). Isomap keeps the distance relationship between samples after dimensionality reduction unchanged, and find the real manifold of high-dimensional data so that the results after dimensionality reduction can more accurately correspond to high-dimensional feature vectors. Therefore, we choose Isomap as the dimensionality reduction method used by the system.

\section{EVALUATION}

\subsection{Performance Metrics}

In order to accurately evaluate the effectiveness of our method, we choose the visualization results as the subjective evaluation metrics, and at the same time, we choose the performance metrics of the network as the objective evaluation metrics.

\textbf{Subjective evaluation metrics.}
The visualization results of retraining are an important consideration in evaluating our method. The visualization results can not only reflect the quality of the network classification results but also reflect whether the user's editing of the latent space is effective. If the user-edited retrained projection results are close to the user-edited results, we can assume that the network has learned the user-edited latent space.

\textbf{Objective evaluation metrics.}
We use micro-F1 and ROC commonly used in the classification field as performance metrics to evaluate the performance of the network. For the final retraining results, we apply ROC for evaluation, and at the same time, we plot the micro-F1 of each epoch as a micro-F1 curve.


\subsection{Datasets Selection}

\begin{figure}[h]
\centering
\includegraphics[width=0.45\textwidth]{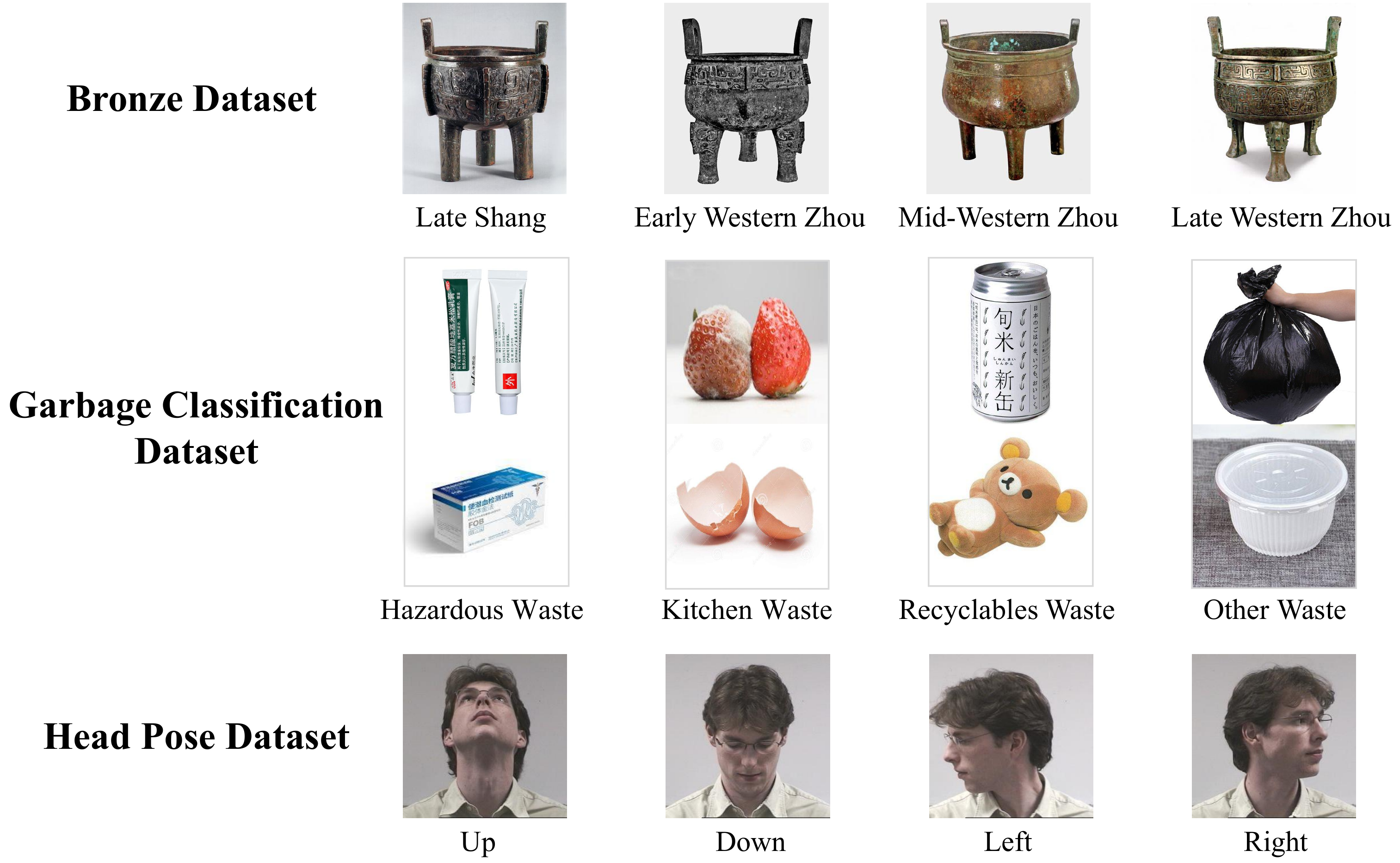}
\renewcommand*{\thefigure}{9}
\caption{\label{fig:figuredataset} Example images for the three selected datasets.}
\end{figure}

We evaluate the proposed system using three datasets of varying difficulty, namely bronze dataset (self-collected), 
garbage classification dataset \cite{rrubbish}, and head pose dataset \cite{rfacepose} (Fig. \ref{fig:figuredataset}). And we invited users of different identities to conduct case studies on these datasets. 
Each of the three types of datasets selected for the experiments has four classes. The bronze dataset for experiments consists of 800 images, of which about 400 are train set, about 150 are validation set, and about 250 are test set. Both the garbage classification dataset and the head pose dataset for experiments have 700 images, of which about 400 are train set, about 100 are validation set, and the remaining about 200 are test set. The experimental results will be used to demonstrate whether our designed system is useful in different scenarios.

The bronze dataset represents datasets in the specialized field, and the similarities between bronzes from adjacent ages are so great that only experienced archaeologists can accurately classify bronzes. There are multiple sub-classes under each class in the garbage classification dataset, which poses a certain challenge to the classification effect of the network. The head pose dataset has orientation properties that are difficult for networks to discern, and networks are particularly poor at distinguishing facing left from facing right. When it comes to discriminating images in the three datasets above, humans have an advantage that machines cannot.

















\section{CASE STUDY}

\subsection{Approach}
We first train on raw data to obtain a network for visualizing images. Then, the high-dimensional features are projected into a 2D workspace through the network, which facilitates user interaction with high-dimensional hidden vectors. The user can help network learning by adjusting the position of projected points in the 2D workspace. When the user operates the projection points on the 2D workspace, the network will learn according to the changes before and after the projection points, therefore achieving the purpose of artificially assisted network retraining.

\subsection{Task}

In order to verify the function of the system and the effectiveness of the method, we invited users of different identities to evaluate. Their task is to move the data in the 2D workspace based on their knowledge, then retrain the network and evaluate the updated results. The conditions for stopping the update are at the user's discretion. The process of conducting a case study is as follows:

\begin{enumerate}
    \item \textbf{Preparation stage.} Users watch a video explaining the principles and functions of our system, after which we further explain to users how the system works and their tasks, and answer their questions;
    \item \textbf{Practice stage.} Let users use the system for 10 minutes to familiarize themselves with system functions;
    \item \textbf{Test stage.} The user officially started the test. For the training sets of different difficulty, we stipulated different operation times, including 45 minutes for the bronze dataset, 25 minutes for the garbage classification dataset, and 20 minutes for the head pose dataset;
    \item \textbf{Interview stage.} Interview with the user about the functionality of the system and the results of the case study.
\end{enumerate}

\subsection{Bronze Dataset}
The difficulty in dating bronzes lies in whether the judges can grasp the characteristics of each age. At the same time, they are also required to understand the styles, crafts, and casting methods formed in different regions. Often only well-trained experts can make accurate judgments, and experts also need to compare the features between images many times when making judgments. In this case study, we invited a researcher in the field of archaeology, \begin{math}\mathit{E}\end{math}$_{A}$, to use our system, who had seven years of studying archaeology experience and had participated in the task of collation and classification of large datasets of bronzes.

\subsubsection{User Operation}

We prepared the dataset and pretrained network in advance for \begin{math}\mathit{E}\end{math}$_{A}$ as needed. \begin{math}\mathit{E}\end{math}$_{A}$ first looked at the classes of the dataset and found that there were bronzes from the early Western Zhou and late Shang dynasties in the data. Based on his expertise, he speculates that the classification accuracy of the data from these two dynasties may be lower. Then, he carefully observed the distribution of data projections and found that there was no clear boundary between the data projections corresponding to these two dynasties, which confirmed his conjecture. At the same time, he found that the data far from the center of the class cluster is often located in the transition period of age. This part of the data is often mixed with the common characteristics of the two ages. The classification results of the network on this part of the data are not very good, therefore he focused on this part of the data. Then, he takes the center position of each class cluster as the reference frame, and the data far from the center of the cluster as the main mobile data. His method is to classify the data at the cluster center according to a certain characteristic of bronzes, such as shape, therefore he divides the data at the center of each class cluster into several sub-classes. After that, he moves the data away from the cluster center to where he sees fit in the cluster center. He said in a later interview that he wanted to subdivide the interior of each class based on important features such as the shape and decoration of the bronzes. Through the interactive features designed by the system, the researcher can easily move the data where he seems reasonable.

\begin{figure*}
\centering
\includegraphics[width=1.0\textwidth]{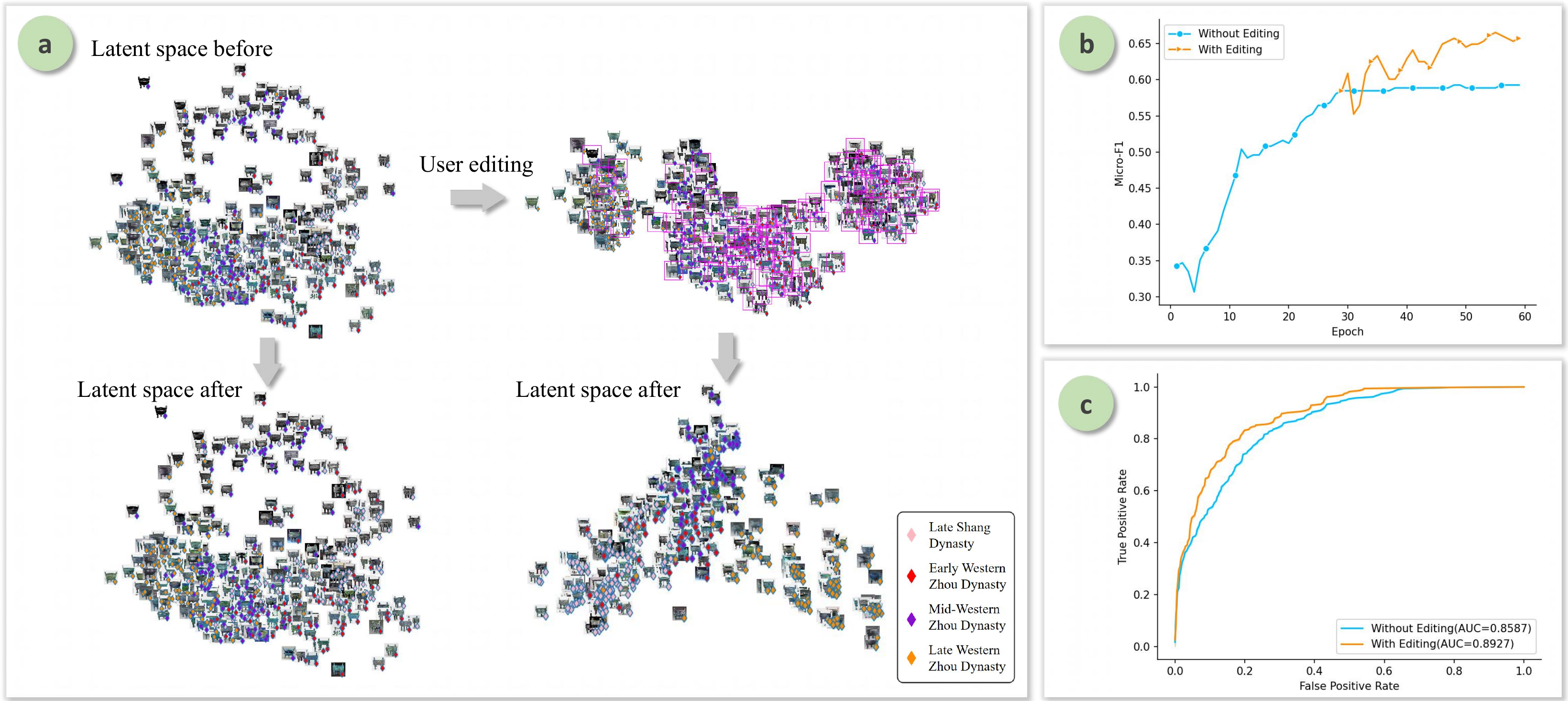}
\renewcommand*{\thefigure}{10}
\caption{\label{fig:figure14} 
Study results on the bronze dataset. (a) Comparison of retrained projection distributions without and with user editing. (b) Comparison of micro-F1 curves obtained by retraining without and with user editing. (c) Comparison of ROC curves obtained from the final results of retraining without and with user editing.
}
\end{figure*}

\subsubsection{Results}
There is no clear boundary between the retrained projection distribution without \begin{math}\mathit{E}\end{math}$_{A}$ editing, but sharp boundaries emerged between the retrained projection distribution with \begin{math}\mathit{E}\end{math}$_{A}$ editing (Fig. \ref{fig:figure14}(a)). Interestingly, we also find that the retrained projection distribution with \begin{math}\mathit{E}\end{math}$_{A}$ editing from left to right corresponds to the order of the bronze age. From the results of micro-F1 curves (Fig. \ref{fig:figure14}(b)) and ROC curves (Fig. \ref{fig:figure14}(c)), it can be seen that the performance of the network has been significantly improved after \begin{math}\mathit{E}\end{math}$_{A}$ editing.

\subsubsection{User Feedback}



$``$The system can prompt me with similarities and differences between data that I would otherwise ignore.$"$ \begin{math}\mathit{E}\end{math}$_{A}$ believes that the system can help him discover relationships between data, which is ideal for his use case: partitioning data by constantly comparing the differences between them. He can easily observe the data using the capabilities of the system, and he can further compare similarities and differences between the data using the history record module. He even comes up with new usages, such as using the spatial layout structure we designed to divide the task of archaeological data, which can greatly reduce his workload. In addition, he also commented, $``$Both shape and decoration are very important to the classification of bronzes, and it is difficult for us to take into account both when we manually classify these data, but the use of 2D layout structure solves this problem to a certain extent.$"$ Placing data with multiple characteristics between clusters, rather than just one of them, is an advantage of a 2D spatial layout.

However, \begin{math}\mathit{E}\end{math}$_{A}$ also said that people in the field of archaeology often use various software to process the data. From his point of view, as a person who is not in the field of deep learning, it is still difficult for our system to get started, and a certain adaptation stage is required. At the same time, archaeological datasets are inherently difficult, so more straightforward systems tend to be more applicable.

\subsection{Garbage Classification Dataset}
The difficulty of garbage classification datasets is that there are many subclasses under each class. We invited deep learning researcher \begin{math}\mathit{E}\end{math}$_{B}$ to conduct this case study, he has 7 years of computer learning experience and 4 years of deep learning experience, and his main research direction is fine-grained classification. At the same time, he is proficient in basic garbage sorting rules.

\begin{figure*}
\centering
\includegraphics[width=1.0\textwidth]{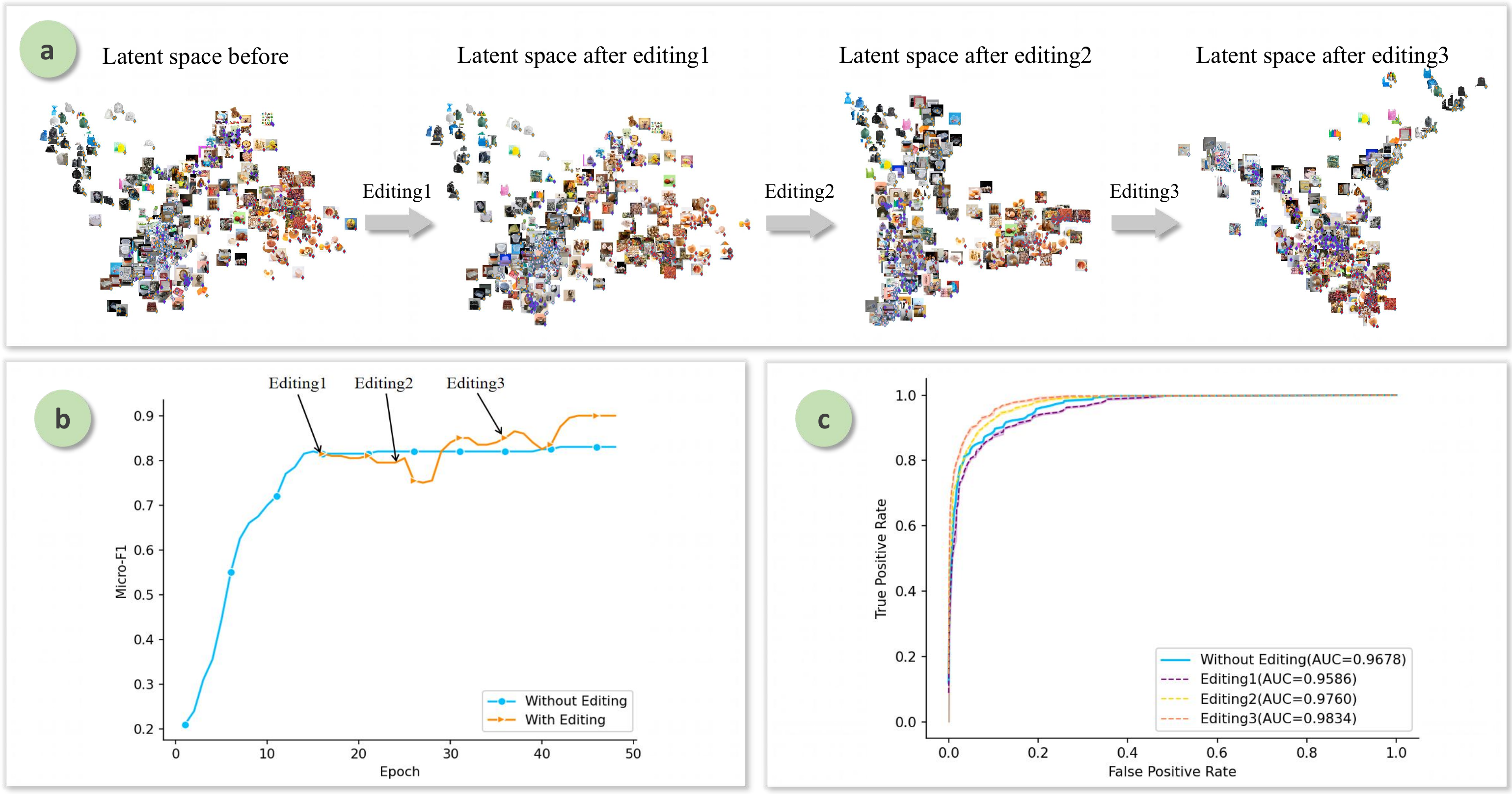}
\renewcommand*{\thefigure}{11}
\caption{\label{fig:figure15} 
Study results on the garbage classification dataset. (a) Comparison of retrained projection distributions without and with three edits by the user. (b) Comparison of micro-F1 curves obtained by retraining without and with three edits by the user. (c) Comparison of ROC curves obtained from the final results of retraining without and with three edits by the user.}
\end{figure*}

\subsubsection{Process}

\begin{math}\mathit{E}\end{math}$_{B}$ initially used the system's adjust display points and zoom in comparison functions. In post-use interviews, \begin{math}\mathit{E}\end{math}$_{B}$ said he had experience designing user interfaces, therefore he paid particular attention to whether the system could clearly display data and compare data easily. \begin{math}\mathit{E}\end{math}$_{B}$ believes that the functions designed by the system are very comprehensive, and the logic between each component is smooth, which can meet his needs.

\begin{math}\mathit{E}\end{math}$_{B}$ compared the relationship between the data in combination with our system and current classification accuracy. He found that classes with low classification accuracy were projected less densely, while classes with high classification accuracy were projected more tightly. He speculates that the reason may be that the network's feature learning of this part of the data is not thorough enough. Therefore, he focuses on predicting data that is not dense enough. He first moved a small amount of deviated data to the vicinity of the corresponding cluster, and then set the training period to 8 epochs. But the result after retraining did not change much. He guessed it might be because he was moving too little data. From the perspective of the relationship between the fully connected layer and the latent space, he intends to move the data with wrong predicted values to the vicinity of the data with correct predicted values, and at the same time perform intra-class aggregation operations on the data of each class. In a subsequent interview, he stated that the fully connected layer of the network can be seen as a function, therefore whether the predicted value is correct depends on the independent variable features. If the wrongly predicted features are closer to the correctly predicted features, the output of the network is also more likely to change from wrongly predicted to correctly predicted. Then, \begin{math}\mathit{E}\end{math}$_{B}$ made a second retraining according to the above idea, this time the training period is 12 epochs. The accuracy of this training has been significantly improved. At the same time, \begin{math}\mathit{E}\end{math}$_{B}$ believes that there is a certain gap between the classes in the visualization results after retraining, so he believes that his idea is feasible. Then, he continued the above idea, focusing the third time on the purple class with no clear boundaries. He mainly separated the purple class from the others and then did a third retraining. After the third retraining, the accuracy rate improved significantly. At the same time, there are obvious gaps between the classes in the visualization results. \begin{math}\mathit{E}\end{math}$_{B}$ is very satisfied with the result of this movement, and then he ends the operation.



\subsubsection{Results}
Through the retrained visualization results without editing and with three editing by \begin{math}\mathit{E}\end{math}$_{B}$ (Fig. \ref{fig:figure15}(a)), we can clearly feel the process of separating the originally mixed data. We can also feel the changes brought by user interaction to the network from the micro-F1 curves (Fig. \ref{fig:figure15}(b)) and ROC curves (Fig. \ref{fig:figure15}(c)). At the same time, the metric curves and visualization results demonstrate the effectiveness of our method.

\subsubsection{User Feedback}

\begin{math}\mathit{E}\end{math}$_{B}$ praised the utility of the system and commented, $``$The system is so novel that it is intelligible to be able to connect the latent space with data so directly.$"$ At the same time, \begin{math}\mathit{E}\end{math}$_{B}$ also said that he is more concerned about how the system presents data to the user. He believes that the system we designed is fully functional, not only has basic zooming, viewing, and guiding functions, but also can adjust the display density, which can display data well, and the system is also very user-friendly. In addition, \begin{math}\mathit{E}\end{math}$_{B}$ also repeatedly observed the results of retraining. He believes that the network is gradually learning the latent space after his movement, and the retrained network can also show better performance.

\begin{math}\mathit{E}\end{math}$_{B}$ also suggested to the system that a 2D spatial layout might not be sufficient when dealing with complex distributions of data points. Although we use Isomap, a dimensionality reduction method that can correspond to low-dimensional and high-dimensional, it cannot fully reflect the relationship between features. In future work, we will try to elevate the spatial layout to three-dimensional space.



\begin{figure*}
\centering
\includegraphics[width=1.0\textwidth]{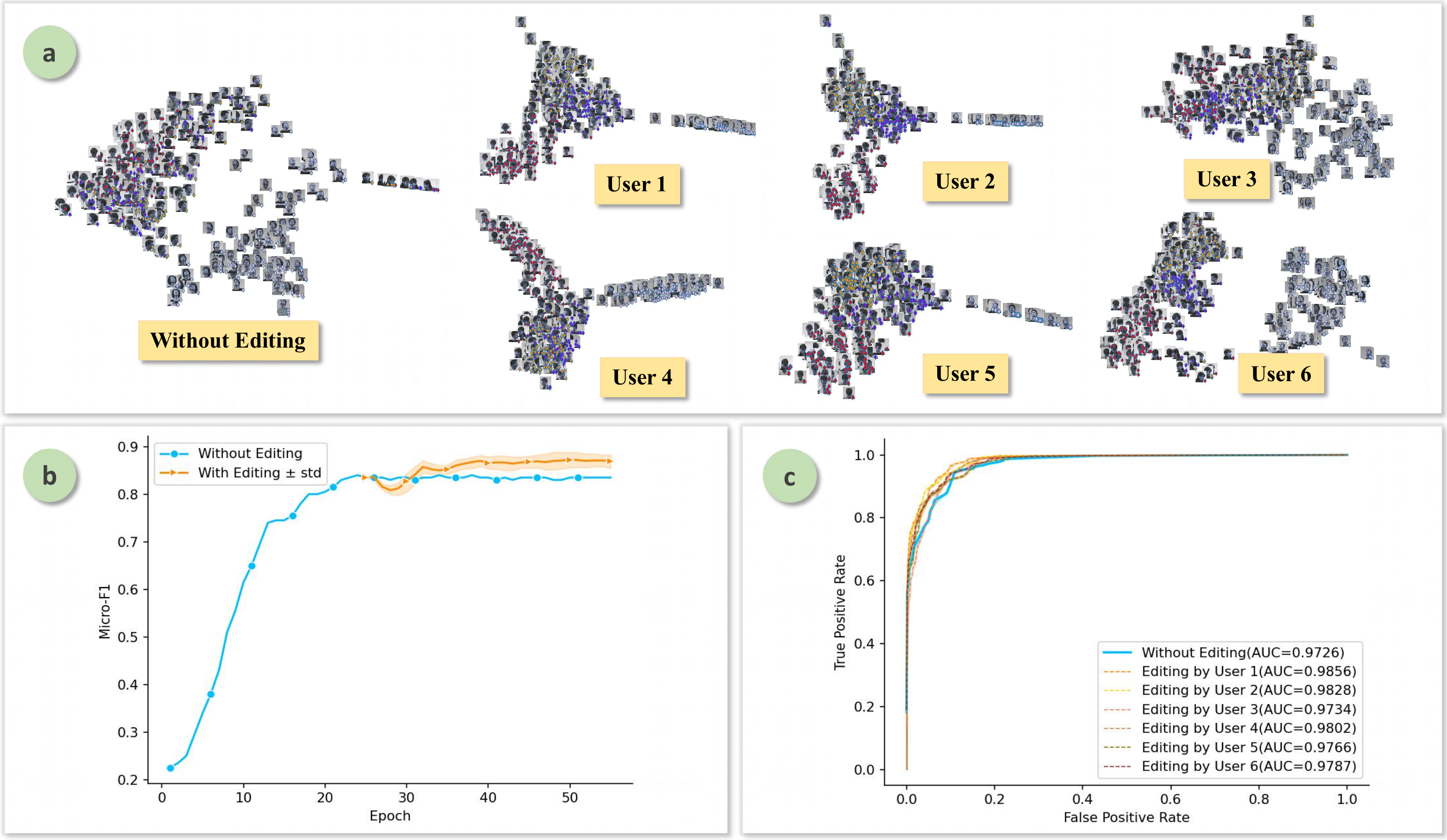}
\renewcommand*{\thefigure}{12}
\caption{\label{fig:figure16} 
Study results on the head pose dataset. (a) Comparison of retrained projection distributions without and with editing by six users. (b) Comparison of mean micro-F1 curves obtained by retraining without and with editing by six users. (c) Comparison of ROC curves obtained from the final results of retraining without and with editing by six users.}
\end{figure*}

\subsection{Head Pose Dataset}
The head pose dataset includes four classes, facing up, facing down, facing left, and facing right. We invited 6 graduate students (3 males, 3 females) to join our system evaluation, their majors include bioinformatics, computer vision, and recommender systems, and all of them are deep learning beginners.

\subsubsection{Process}
After simple exercises, they can quickly master the skills of using the system and understand the meaning of the spatial layout of the system.

They generally check the classification of the projection data from the beginning. Four users found that the projection results of the facing up class have the highest degree of aggregation, and the facing up class has the highest classification accuracy. They feel that the density of the projection data will affect the accuracy. They believe that aggregating scattered data can be an efficient way to move. Another user found that the projected data for the facing left and facing right classes were almost mixed, with the lowest classification accuracy for both. Combining his knowledge, he assumed that neural networks are not very distinguishable between the concepts of the facing left and facing right classes, therefore he focused on dealing with facing left and facing right classes. The last user knows some visualization techniques, so she pays more attention to the distribution of projections before and after moving. She observes that the data for the four classes are more or less mixed together. $``$If I separate the data from the four classes that are mixed, will the network learn such a distribution?$"$ With this idea in mind, she applies the functions of the system to process the projection data.

\subsubsection{Results}
Compared to the without editing visualization result, the six user visualization results produced clear demarcations between all four classes (Fig. \ref{fig:figure16}(a)). In particular, the three classes of purple, red, and orange that were originally mixed can have a clear cluster structure after editing and retraining by the users. At the same time, it can be seen from the results of micro-F1 curves (Fig. \ref{fig:figure16}(b)) and ROC curves (Fig. \ref{fig:figure16}(c)) that the users' editing has significantly improved the performance of the network.

\subsubsection{User Feedback}



All six users expressed a positive view of the retrained results, they all felt that the updated network learned their $``$knowledge$"$ and that by moving the data process and results, they were able to further understand the concept of latent space. One of the users commented, $``$The method used by this system meets my needs in an application, and it solves many invisible problems encountered in network training, making the whole process more intuitive and clear.$"$ Making the originally uncontrollable training process controllable has always been the direction of our efforts. When it comes to the design of the system interface, they generally believe that such a spatial layout allows them to grasp the spatial location information of the samples, thumbnail images, and the overall distribution of the data, which is concise and intuitive.


\section{Discussion}
\textbf{The system facilitates the observation of data and helps to discover relationships between data.}
The system can intuitively observe the similarities and differences between data, and the user can use the spatial layout of the system to dig out the similarities between adjacent data that may be overlooked, and can also dig out the differences between data that are far away. In addition, the system can help users organize data. For example, archaeologists need to process each image when classifying archaeological data, which is a tedious and tedious process. But by borrowing our spatial layout, they can divide the data according to the projection results, which greatly reduces the workload.

\textbf{The system facilitates the partitioning of data.}
The combination of spatial layout and interactive functions is not only beneficial to classify data between classes but also to classify data within the same class according to attributes such as shape or pattern. At the same time, such a spatial layout can also provide a buffer for the user. For example, due to the continuity of the ages, bronze wares often show the characteristics of two ages. Simply dividing this type of data into a specific age will be biased. Using the two-dimensional layout of our system, the user can solve this problem by placing this part of the data between two age clusters.

\textbf{This system allows users to obtain a more understandable latent space.}
The system connects a high-dimensional latent space with a 2D workspace, allowing users to apply their knowledge to the training process of the network, which can make the network jump out of the local minimum area and improve the performance of the network. At the same time, the original concept of latent space is relatively abstract, but users can intuitively feel the changing process of latent space in the process of using the system we designed.

\textbf{This system allows users to speed up the learning process of the network.}
By visualizing the changes before and after retraining, the user can see that the updated results are closer to the results they moved. Meanwhile, the training process of the original network is unknown, but our system exposes this process to some extent.

\textbf{The system functions are well-designed, easy to use, and can be displayed clearly with data. }
In the latter two case studies, they both rated the system as easy to use, an assessment that differed from the researcher \begin{math}\mathit{E}\end{math}$_{A}$. We can think that for users with a certain deep learning foundation, the system is relatively easy to use. In addition, all users believed that the interactive function settings of the system were very reasonable, the designed functions could meet their needs in use, and the combination of system layout and functions could clearly display the data.

\section{CONCLUSION}

In this paper, we introduce an interactive system that not only connects a high-dimensional latent space with a 2D workspace but also allows the user to interact with the visualized data and retrain the network, enabling humans to help the network learn. The results of three case studies prove that latent vectors can be edited by the user, and through the system we designed, human knowledge in classification can be incorporated into the training process of the network, thereby improving the performance of the network.

\section*{Acknowledgments}
This work is supported in part by the Young Scientists Fund of the National Natural Science Foundation of China (Grant No.62206106) and Jilin University (Grant No.419021422B08).

\bibliographystyle{abbrv-doi}
\bibliography{sample-base}

\end{document}